\documentclass[11pt,a4paper]{article}
\usepackage[hyperref]{acl2021}
\usepackage{times}
\usepackage{latexsym}
\usepackage{graphicx}
\usepackage{float}
\usepackage{todonotes}
\usepackage{multirow}

\usepackage{amsmath}
\usepackage[separate-uncertainty = true, multi-part-units = repeat]{siunitx}

\usepackage{microtype}

\aclfinalcopy 

\title{LXMERT Model Compression for Visual Question Answering}
\author{
Maryam Hashemi \hspace{0.2cm} Ghazaleh Mahmoudi  \Thanks{ These authors contributed equally.}$^*$ \hspace{0.2cm} Sara Kodeiri$^*$ \hspace{0.2cm} Hadi Sheikhi$^*$\hspace{0.2cm} Sauleh Eetemadi  \\
School of Computer Engineering,
Iran University of Science and Technology, Iran \\
\small{\texttt{\{m\_hashemi94, gh\_mahmoodi, sara\_kodeiri, ha\_sheikhi\}@comp.iust.ac.ir, sauleh@iust.ac.ir }}}
\begin{document}
\maketitle
\begin{abstract}
Large-scale pretrained models such as LXMERT are becoming popular for learning cross-modal representations on text-image pairs for vision-language tasks. According to the lottery ticket hypothesis, NLP and computer vision models contain smaller subnetworks capable of being trained in isolation to full performance. In this paper, we combine these observations to evaluate whether such trainable subnetworks exist in LXMERT when fine-tuned on the VQA task. In addition, we perform a model size cost-benefit analysis by investigating how much pruning can be done without significant loss in accuracy. Our experiment results demonstrate that LXMERT can be effectively pruned by 40\%-60\% in size with 3\% loss in accuracy. 
\end{abstract}

\section{Introduction and Related Work}
Over the past few years, many single-modal pretrained models have been proposed. Inspired by this, the vision-and-language pretraining seeks to learn joint representations using visual and textual content to improve the efficiency of vision-language tasks.

Both single-modality and cross-modality pretrained models often have hundreds of millions of parameters. Unfortunately, training these overparametrized models can be prohibitively time-consuming and costly, making them impractical for resource-limited devices. However, cross-modality pretrained models suffer more from the increased model size due to the higher input space dimension. With the task of Visual Question Answering (VQA) \cite{Antol2015} in mind, and its ultimate goal of being helpful to the visually impaired, decreasing V+L model size makes it feasible to use them in limited-resource devices.

To address this problem, model compression techniques such as pruning have been developed. Deep Learning recently enjoyed welcoming a new powerful pruning method: The Lottery Ticket Hypothesis (LTH) \cite{frankle2018the}. LTH has been shown great success in various fields. It could be a powerful tool to understand the parameter redundancy in the current pretrained V+L models. Thus, we aim to apply LTH to LXMERT\cite{Tan2020}, one of the best-performing two-stream V+L models, to fill this gap. We evaluate our work on VQA \cite{Antol2015} and compare it with DistillVLM\cite{Fang_2021_ICCV}, which leverages the knowledge distillation technique to compress large visual-linguistic models.

Similar to this work, \citet{gan2021playing} study LTH for UNITER\cite{chen2020uniter}. However, UNITER is a single stream V+L model, and LXMERT is a two-stream model; our results are consistent with theirs.

\section{Methodology}
In this section, we briefly explain the LXMERT architecture and LTH. Then, we describe how we use LTH to compress the pretrained LXMERT model.

\begin{table*}[ht]
    \centering
    \resizebox{\textwidth}{!}{
    \begin{tabular}{ccccccccccc}
        \hline
        \multicolumn{2}{c}{\multirow{2}{*}{Method}}& \multicolumn{4}{c}{test-dev} &  & \multicolumn{4}{c}{test-std}\\
        \cline{3-6} \cline{8-11}
        \multicolumn{2}{l}{}& Yes/No & Number & other & Overall & & Yes/No & Number & other & Overall\\
        \hline
        \multicolumn{2}{l}{DistillVLM} & - & - & - & 69.6 & & - & - & - & 69.8  \\
        \hline
        \multicolumn{2}{l}{LXMERT} & 88.24 & 54.45 & 63.05 & 72.45 & & 88.29 & 54.37 & 63.18 & 72.63  \\
        \multicolumn{2}{l}{LXMERT (low-magnitude)} & 86.95 $\pm$ 0.95 & 52.60 $\pm$ 1.87 & 60.96 $\pm$ 1.76 & 70.72 $\pm$ 1.44 & & 87.07 $\pm$ 1.12 & 52.28 $\pm$ 1.66 & 61.02 $\pm$ 1.83 & 70.87 $\pm$ 1.51  \\
        \multicolumn{2}{l}{LXMERT (high-magnitude)} & 74.11 $\pm$ 0.91 & 42.81 $\pm$ 1.36 & 50.5 $\pm$ 0.19 & 59.35 $\pm$ 0.61 & & 74.23 $\pm$ 0.81 & 42.99 $\pm$  0.87 & 50.71 $\pm$ 0.26 & 59.62 $\pm$ 0.55  \\
        \multicolumn{2}{l}{LXMERT (random)} & 69.26 $\pm$ 0.29 & 39.84 $\pm$ 0.93 & 45.96 $\pm$ 0.83 & 54.86 $\pm$ 0.52 & & 69.27 $\pm$ 0.18 & 40.34 $\pm$ 0.66 & 46.33 $\pm$ 0.79 & 55.19  $\pm$ 0.45  \\
        \hline
    \end{tabular}
    }
    \caption{\label{table1} Performance of subnetworks at 50\% weights pruning on VQA v2, which reported for both test-dev and test-std. Test-dev is used for debugging and validation experiments. Test-standard is the default test data for the VQA competition. We test each experiment for three different seeds and report the mean and standard deviation of VQA accuracy across three seeds.}
    \vspace{-2mm}    
\end{table*}

LXMERT is a Transformer-based model which takes two inputs: image and text. Internally, LXMERT consists of two types of encoders: single-modality encoders for each modality and a cross-modality encoder using bidirectional cross attention to exchange information and align entities across the modalities.

The Lottery Ticket Hypothesis \cite{frankle2018the} shows that by preserving the original weight initializations from the unpruned network, you can train a network with the topology of the pruned network and achieve the same or better test accuracy within the same number of training iterations.

In order to apply LTH to the LXMERT model, we use iterative magnitude pruning. Therefore, we fine-tune LXMERT on the VQA task and iteratively prune 10\% of the lowest magnitude weights across the entire model, excluding embedding and output layers. We keep pruning until our model loses roughly half the weights. We use the default settings and hyperparameters of LXMERT \cite{Tan2020} to finetune on the VQA v2.0 dataset.

\section{Experimental Setups and Results}

The experiments are designed to investigate the effectiveness and stability of LTH on LXMERT in addition to cost-benefit analysis of the number of parameters in the model. We conduct experiments on the widely-used VQA v2.0 \cite{goyal2017making} dataset built based on the MS-COCO \cite{lin2014microsoft} image corpus.

\subsection{Effectiveness and Stability}
The following steps are performed to compress the LXMERT model.
\begin{enumerate}
    \itemsep0em
    \item The pretrained LXMERT model plus the VQA classifier's randomly initialized weights are saved.
    \item The model is fine-tuned on the 3,129 most frequent answers in the VQA v2.0 dataset.
    \item Iterative magnitude pruning is applied to find the low-magnitude subnetwork (pruning 50\% of the low-magnitude weights). The high-magnitude subnetwork is computed as a compliment of the low-magnitude subnetwork with equal size. A random subnetwork with an equal size is generated for comparison.
    \item The saved weights are restored for all three subnetworks.
    \item The high-magnitude, low-magnitude, and random subnetworks are fine-tuned and evaluated on the VQA Task using three different seeds for initializing the VQA model to ensure the stability of the results.
\end{enumerate}

Results of subnetworks at 50\% weights pruning on VQA v2.0 are summarized in Table \ref{table1} where DistillVLM \cite{Fang_2021_ICCV} is also listed for comparison. Row 2 to row 5 reports respectively full finetuned LXMERT, low-magnitude, high-magnitude, and random subnetworks. Low-magnitude pruning achieves 97\% of full finetuned LXMERT accuracy in overall for both test-dev and test-std and shows marginal improvement over DistillVLM as the baseline. By comparing performance across the subnetworks, random and high magnitude subnetworks perform far worse than low-magnitude subnetwork. Surprisingly, the results demonstrate high-magnitude subnetwork performing better than random subnetwork. This could be a LXMERT specific phenomenon and required further investigation.

\subsection{Cost-Benefit Analysis}
\begin{figure}[h!]
	\centering\includegraphics[width=0.4\textwidth]{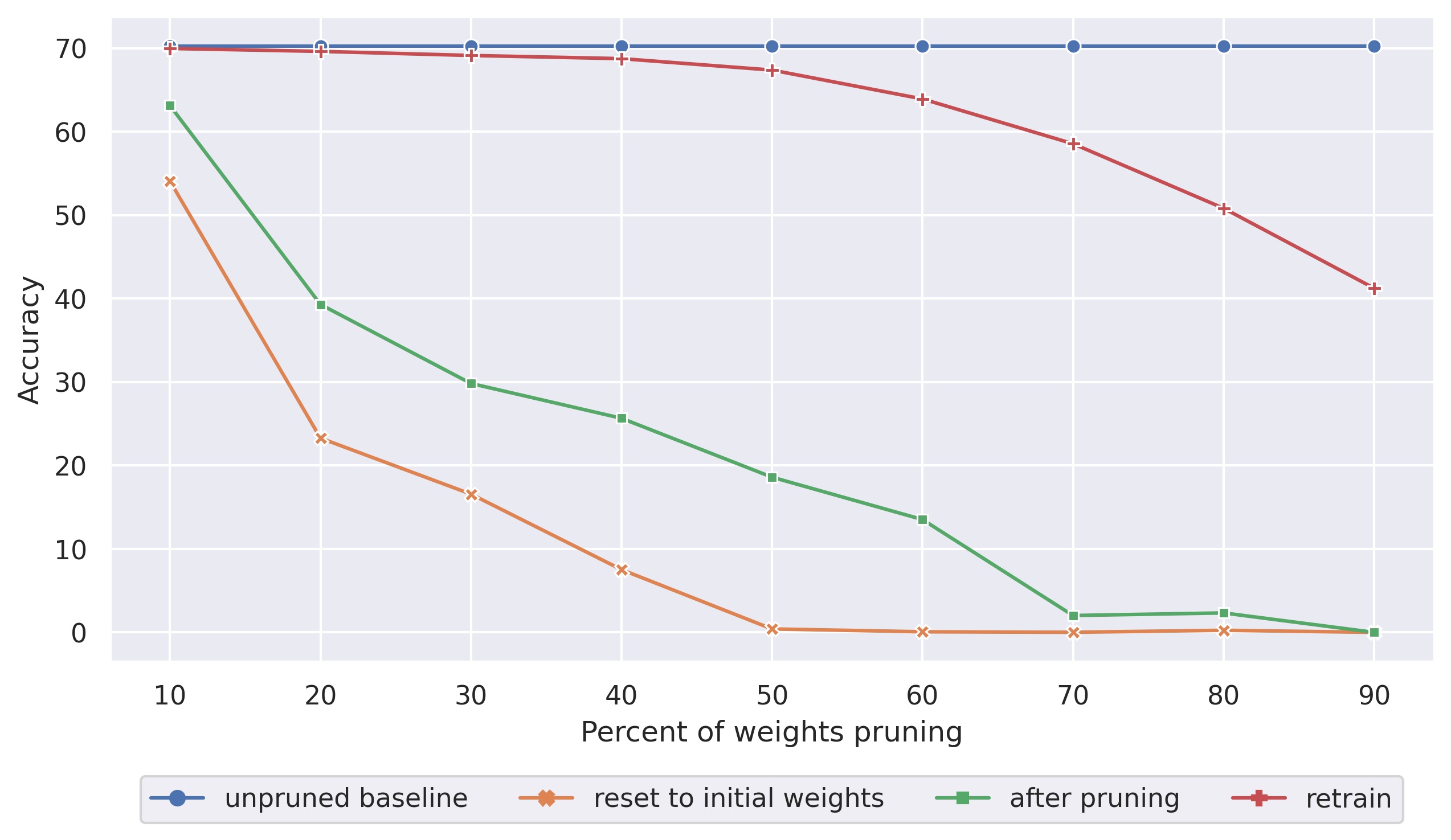}
	\caption{Model size cost-benefit analysis.}
	\label{costbenefit}
    \vspace{-2mm}    	
\end{figure}

We experiment with low-magnitude subnetwork by pruning 10\% of the weights all the way up to 90\% of the weights in 10\% increments. Accuracy of these pruned models on VQA v2.0 are reported in Figure \ref{costbenefit}. Our results indicate a significant loss of accuracy after 50\% to 60\% pruning. 

\section{Conclusion}
We confirm that LTH pruning is an effective method for pruning V+L pretrained models. We mainly focused on LXMERT, a two-stream V+L pretrained model, but our findings are consistent with \citet{gan2021playing}'s results while using UNITER, a single-stream V+L pretrained model.  

\bibliographystyle{acl_natbib}
\bibliography{acl2021}
\end{document}